\def\year{2020}\relax
\documentclass[letterpaper]{article} 
\usepackage{aaai20}  
\usepackage{times}  
\usepackage{helvet} 
\usepackage{courier}  
\usepackage[hyphens]{url}  
\usepackage{graphicx} 
\usepackage{graphicx}  

\usepackage{epsfig}
\usepackage{amsmath}
\usepackage{amssymb}
\usepackage{soul}
\usepackage[utf8]{inputenc}
\usepackage[small]{caption}
\usepackage{booktabs}
\usepackage[ruled,lined,linesnumbered]{algorithm2e}
\usepackage{algorithmic}
\usepackage{xcolor}
\usepackage{adjustbox}
\usepackage{subfigure}
\usepackage{blindtext}
\usepackage{enumerate}
\usepackage{bm}
\usepackage{dsfont}
\usepackage{multirow}
\usepackage[switch]{lineno}

\title{An End-to-End Visual-Audio Attention Network for Emotion Recognition \\in User-Generated Videos}
\author{
Sicheng Zhao$^{1\#}$, Yunsheng Ma$^{23\#}$, Yang Gu$^{2}$, Jufeng Yang$^{4}$\thanks{Corresponding Author. $^{\#}$ Equal Contribution.}, \\
\Large \textbf{Tengfei Xing$^{2}$, Pengfei Xu$^{2}$, Runbo Hu$^{2}$, Hua Chai$^{2}$, Kurt Keutzer$^{1}$}\\
$^{1}$University of California, Berkeley, USA  $^{2}$Didi Chuxing, China\\
  $^{3}$Harbin Institute of Technology, Weihai, China  $^{4}$Nankai University, China\\
  schzhao@gmail.com, 
  yunsheng.ma98@gmail.com, yangjufeng@nankai.edu.cn\\
  \{guyangdavid,xingtengfei,xupengfeipf,hurunbo,chaihua\}@didiglobal.com, keutzer@berkeley.edu \\
}

\begin{document}

\maketitle

\begin{abstract}
Emotion recognition in user-generated videos plays an important role in human-centered computing. Existing methods mainly employ traditional two-stage shallow pipeline, \textit{i.e.} extracting visual and/or audio features and training classifiers. In this paper, we propose to recognize video emotions in an end-to-end manner based on convolutional neural networks (CNNs). Specifically, we develop a deep Visual-Audio Attention Network (VAANet), a novel architecture that integrates spatial, channel-wise, and temporal attentions into a visual 3D CNN and temporal attentions into an audio 2D CNN. Further, we design a special classification loss, \textit{i.e.} polarity-consistent cross-entropy loss, based on the polarity-emotion hierarchy constraint to guide the attention generation. Extensive experiments conducted on the challenging VideoEmotion-8 and Ekman-6 datasets demonstrate that the proposed VAANet outperforms the state-of-the-art approaches for video emotion recognition. Our source code is released at: \url{https://github.com/maysonma/VAANet}.
\end{abstract}


\section{Introduction}
\label{sec:Introduction}

The convenience of mobile devices and social networks has enabled users to generate videos and upload to Internet in daily life to share their experiences and express personal opinions. As a result, an explosive growing volume of videos are being created, which results in urgent demand for the analysis and management of these videos. Besides the objective content recognition, such as objects and actions~\cite{zhu2018towards,choutas2018potion}, understanding the emotional impact of the videos plays an important role in human-centered computing. On the one hand, the videos can, to a large extent, reflect the psychological states of the video generators. We can predict the generators' possible extreme behaviors, such as depression and suicide, and take corresponding preventive actions. On the other hand, the videos that evoke strong emotions can easily resonate with viewers and bring them immersive watching experiences. Appropriate emotional resonation is crucial in intelligent advertising and video recommendation. Further, emotion recognition in user-generated videos (UGVs) can help companies analyze how customers evaluate their products and assist governments to manage the Internet.

\begin{figure}[t]
   \centering
   \includegraphics[width=1\linewidth]{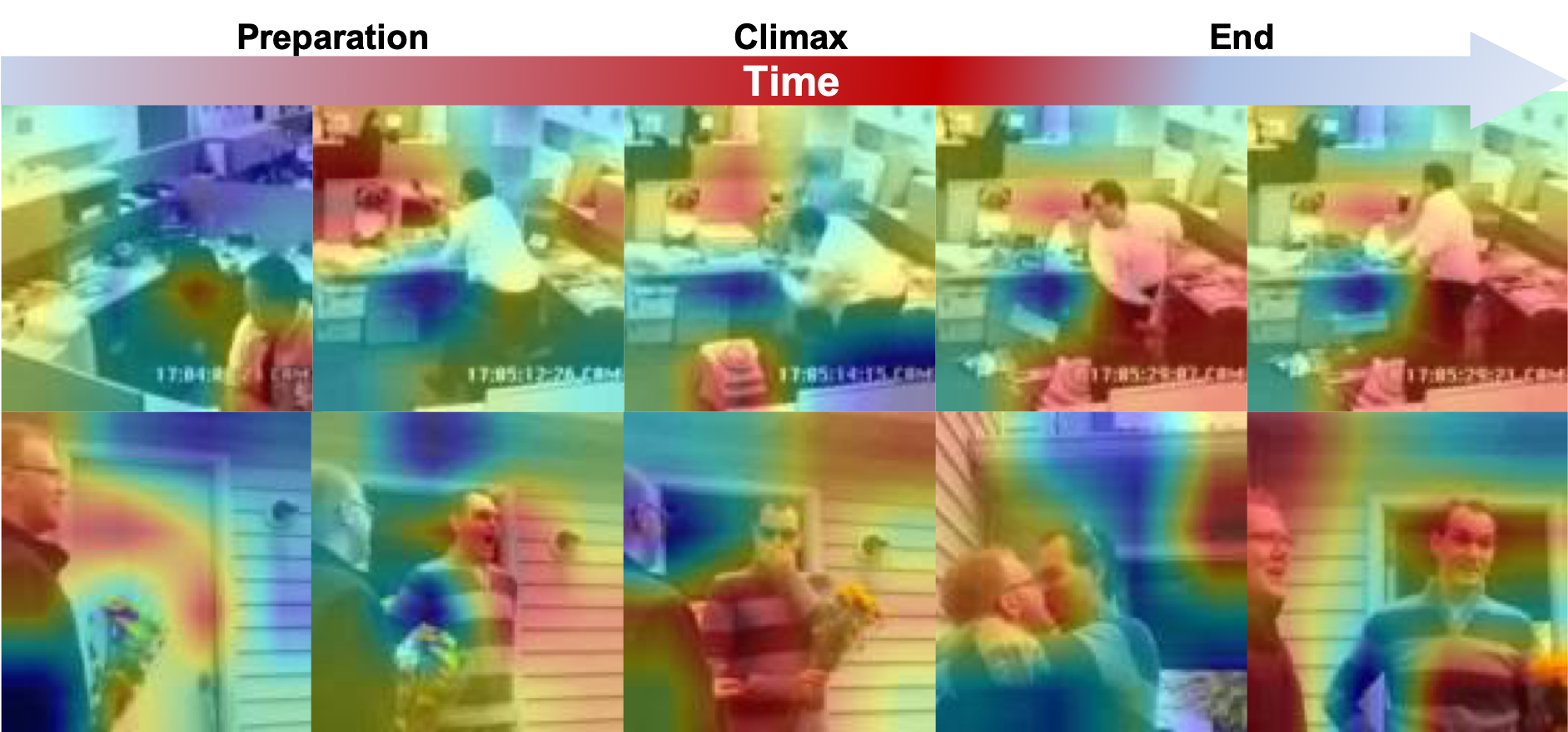}
   \caption{Illustration of the keyframes and discriminative regions for emotion recognition in user-generated videos. Although the story in a video may contain different stages, the emotion is mainly evoked by some keyframes (as shown by the temporal attentions in the color bar) and corresponding discriminative regions (as illustrated by the spatial attentions in the heat map).}
   \label{fig:problem}
\end{figure}

Although with the advent of deep learning, remarkable progress has been made on text sentiment classification~\cite{zhang2018deep}, image emotion analysis~\cite{zhao2018affective,zhao2018predicting,yang2018weakly}, and video semantic understanding~\cite{zhu2018towards,choutas2018potion}. Emotion recognition in UGVs still remains an unsolved problem, due to the following challenges. (1) \textit{Large intra-class variation.} Videos captured in quite different scenarios may evoke similar emotions. For example, visiting an amusement park, taking part in sport competition, and playing video games may all make viewers feel ``excited''. This results in obvious ``affective gap'' between low-level features and high-level emotions. (2) \textit{Low structured consistency.} Unlike professional and commercial videos, such as movies~\cite{wang2006affective} and GIFs \cite{jou2014predicting,yang2019human}, UGVs are usually taken with diverse structures, \textit{e.g.} various resolutions and image blurring noises. (3) \textit{Sparse keyframe expression.} Only limited keyframes directly convey and determine emotions, as shown in Figure~\ref{fig:problem}, while the rest are used to introduce the background and context.

Most existing approaches on emotion recognition in UGVs focus on the first challenge, \textit{i.e.} employing advanced image representations to bridge the affective gap, such as (1) mid-level attribute features~\cite{jiang2014predicting,tu2019multi} like ObjectBank~\cite{li2010object} and SentiBank~\cite{borth2013sentibank}, (2) high-level semantic features~\cite{chen2016emotion} like detected events~\cite{jiang2017exploiting,caba2015activitynet}, objects~\cite{deng2009imagenet}, and scenes~\cite{zhou2014learning}, and (3) deep convolutional neural network (CNN) features~\cite{xu2018heterogeneous,zhang2018recognition}. \citeauthor{zhang2018recognition}~\shortcite{zhang2018recognition} transformed frame-level spatial features to another kernelized feature space via discrete Fourier transform, which partially addresses the second challenge. For the third challenge, the videos are either downsampled averagely to a fixed number of frames~\cite{zhang2018recognition}, or represented by continuous frames from only one segment~\cite{tu2019multi}.

The above methods have contributed to the development of emotion recognition in UGVs, but they still have some problems. (1) They mainly employ a two-stage shallow pipeline, \textit{i.e.} extracting visual and/or audio features and training classifiers. (2) The visual CNN features of each frame are separately extracted, which ignore the temporal correlation of adjacent frames. (3) The fact that emotions may be determined by keyframes from several discrete segments is neglected. (4) Some methods require auxiliary data, which is not always available in real applications. For example, the extracted event, object, and scene features in~\cite{chen2016emotion} are trained on FCVID~\cite{jiang2017exploiting} and ActivityNet~\cite{caba2015activitynet}, ImageNet~\cite{deng2009imagenet}, and Places205~\cite{zhou2014learning} datasets, respectively. (5) They do not consider the correlations of different emotions, such as the polarity-emotion hierarchy constraint, \textit{i.e.} the relation of two different emotions belonging to the same polarity is closer than those from opposite polarities.

In this paper, we propose an end-to-end Visual-Audio Attention Network, termed VAANet, to address the above problems for recognizing the emotions in UGVs, without requiring any auxiliary data except the data for pre-training. First, we spit each video into an equal number of segments. Second, for each segment, we randomly select some successive frames and feed them into a 3D CNN~\cite{hara2018can} with both spatial and channel-wise
attentions to extract visual features. Meanwhile, we transform the corresponding audio waves into spectrograms and feed them into a 2D CNN~\cite{he2016deep} to extract audio features. Finally, the visual and audio features of different segments are weighted by temporal attentions to obtain the whole video's feature representation, which is followed by a fully connected layer to obtain emotion predictions. Considering the polarity-emotion hierarchy constraint, we design a novel classification loss, \textit{i.e.} polarity-consistent cross-entropy (PCCE) loss, to guide the attention generation.


In summary, the contributions of this paper are threefold:

\begin{enumerate}
\item We are the first to study the emotion recognition task in user-generated videos in an end-to-end manner.
\item We develop a novel network architecture, \textit{i.e.} VAANet, that integrates spatial, channel-wise, and temporal attentions into a visual 3D CNN and temporal attentions into an audio 2D CNN for video emotion recognition. We propose a novel PCCE loss, which enables VAANet to generate polarity preserved attention map.
\item We conduct extensive experiments on the VideoEmotion-8~\cite{jiang2014predicting} and Ekman-6~\cite{xu2018heterogeneous} datasets, and the results demonstrate the superiority of the proposed VAANet method, as compared to the state-of-the-art approaches.
\end{enumerate}

\begin{figure*}[!t]
\begin{center}
\centering \includegraphics[width=1.0\linewidth]{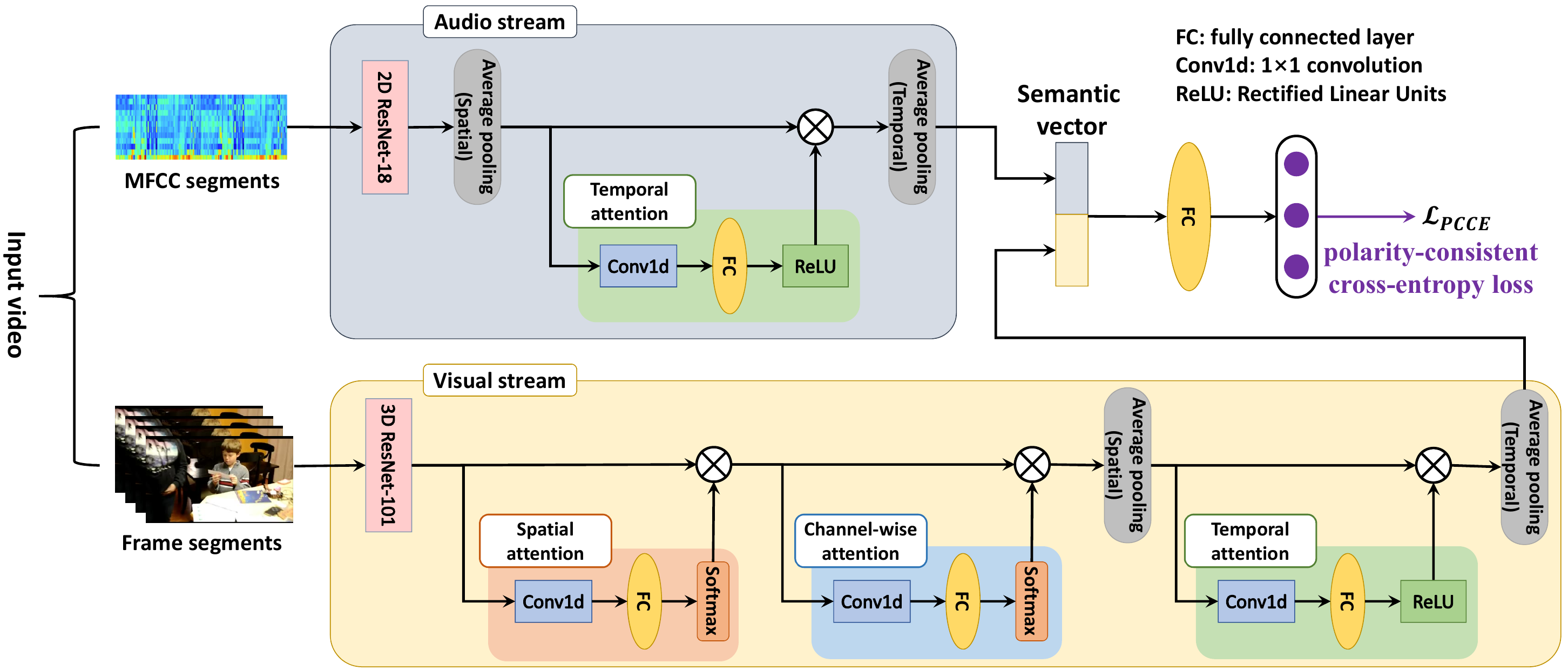}
\caption{The framework of the proposed Visual and Audio Attention Network (VAANet). First, the MFCC descriptor from the soundtrack and the visual information are both divided into segments and fed into 2D ResNet-18 and 3D ResNet-101 respectively to extract audio and visual representation. The response feature maps of the visual stream are then fed into the stacked spatial attention, channel-wise attention, and temporal attention sub-networks, and the response feature map of the audio stream are fed into a temporal attention module. Finally, the attended semantic vectors that carry visual and audio information are concatenated. Meanwhile, a novel polarity-consistent cross-entropy loss is optimized to guide the attention generation for video emotion recognition.}

\label{fig:Framework}
\end{center}
\end{figure*}

\section{Related Work}
\label{sec:RelatedWork}

\noindent\textbf{Video Emotion Recognition:} Psychologists usually employ two kinds of models to represent emotions: categorical emotion states (CES) and dimensional emotions space (DES). CES classify emotions into several basic categories, such as Ekman's 6 basic categories~\cite{ekman1992argument} and Plutchik's wheel of emotions~\cite{plutchik1980emotion}. DES usually employ a Cartesian space to represent emotions, such as valence-arousal-dominance~\cite{schlosberg1954three}. Since CES are easy for users to understand and label, here we adopt CES to represent emotions in videos.

Early research on video emotion recognition mainly focused on movies, which are well structured. \citeauthor{kang2003affective}~\shortcite{kang2003affective} employed a Hidden Markov Model to detect affective event based on low-level features, including color, motion, and shot cut rate. Joint combination of visual and audio features with support vector machine~\cite{wang2006affective} and conditional random fields~\cite{xu2013hierarchical} achieves promising result. Some recent methods work on Animated GIFs~\cite{jou2014predicting,chen2016predicting,yang2019human}. \citeauthor{jou2014predicting}~\shortcite{jou2014predicting} firstly proposed to recognize GIF emotions by using features of different types. \citeauthor{chen2016predicting}~\shortcite{chen2016predicting} improved the performance by adopting 3D ConvNets to extract spatiotemporal features. Human-centered GIF emotion recognition is conducted by considering human related information and visual attention~\cite{yang2019human}.

Because of the content diversity and low quality, UGVs are more challenging to recognize emotions. \citeauthor{jiang2014predicting}~\shortcite{jiang2014predicting} investigated a large set of low-level visual-audio features and mid-level attributes, \textit{e.g.} ObjectBank~\cite{li2010object} and SentiBank~\cite{borth2013sentibank}. \citeauthor{chen2016emotion}~\shortcite{chen2016emotion} extracted various high-level semantic features based on existing detectors. Compared with hand-crafted features, deep features are more widely used~\cite{xu2018heterogeneous,zhang2018recognition}. By combining low-level visual-audio-textual features, \citeauthor{pang2015deep}~\shortcite{pang2015deep} showed that learned joint representations are complementary to hand-crafted features. Different from these methods, which employ a two-stage shallow pipeline, we propose the first end-to-end method to recognize emotions in UGVs by extracting attended visual and audio CNN features.

Please note that emotion recognition has also been widely studied in other modalities, such as text~\cite{zhang2018deep}, images~\cite{zhao2017continuous,yang2018retrieving,zhao2018affective,yang2018weakly,zhao2018emotiongan,zhao2019cycleemotiongan,zhao2019pdanet,yao2019attention,zhan2019zero}, speech~\cite{el2011survey}, physiological signals~\cite{alarcao2017emotions,zhao2019personalized}, and multi-modal data~\cite{soleymani2017survey,zhao2019affective}.

\noindent\textbf{Attention-Based Models:} Since attention can be considered as a dynamic feature extraction mechanism that combines contextual fixations over time~\cite{mnih2014recurrent,chen2017sca}, it has been seamlessly incorporated into deep learning architectures and  achieved outstanding performances in many vision-related tasks, such as image classification~\cite{woo2018cbam}, image captioning~\cite{you2016image,chen2017sca,chen2018show}, and action recognition~\cite{song2017end}. These attention methods can be roughly divided into four categories: spatial attention~\cite{song2017end,woo2018cbam}, semantic attention~\cite{you2016image}, channel-wise attention~\cite{chen2017sca,woo2018cbam}, and temporal attention~\cite{song2017end}.

There are also several methods that employ attention for emotion recognition in images~\cite{you2017visual,yang2018weakly,zhao2019pdanet} and speech~\cite{mirsamadi2017automatic}. The former methods mainly consider spatial attention except PDANet~\cite{zhao2019pdanet} which also employs channel-wise attention, while the latter one only uses temporal attention. To the best of our knowledge, attention has not been studied on emotion recognition in user-generated videos. In this paper, we systematically investigate the influence of different attentions in video emotion recognition, including the importance of local spatial context by spatial attention, the interdependency between different channels by channel-wise attention, and the importance of different segments by temporal attention.

\section{Visual-Audio Attention Network}
\label{sec:VAANet}
We propose a novel CNN architecture with spatial, channel-wise, and temporal attention mechanisms for emotion recognition in user generated videos. Figure~\ref{fig:Framework} shows the overall framework of the proposed VAANet. Specifically, VAANet has two streams to respectively exploit the visual and audio information. The visual stream consists of three attention modules and the audio stream contains a temporal attention module. The spatial attention and the channel-wise attention sub-networks in the visual stream are designed to automatically focus on the regions and channels that carry discriminative information within each feature map. The temporal attention sub-networks in both the visual and audio streams are designed to assign weights to different segments of a video. The training of VAANet is performed by minimizing the newly designed polarity-consistent cross-entropy loss in an end-to-end manner.

\subsection{Visual Representation Extraction}
\label{ssec:Representation}

To extract visual representations from a long-term video, following~\cite{wang2016temporal}, the visual stream of our model works on short snippets sparsely sampled from the entire video. Specifically, we divide each video into $t$ segments with equal duration, and then randomly sample a short snippet of $k$ successive frames from each segment. We use 3D ResNet-101~\cite{hara2018can} as backbone of the visual stream. It takes the $t$ snippets (each has $k$ successive frames) as input and independently processes them up to the last spatiotemporal convolutional layer conv5 into a super-frame. Suppose we are given $N$ training samples $\{(\textbf{x}_l^{V},\textbf{y}_l)\}_{l=1}^{N}$, where $\textbf{x}_l^{V}$ is the visual information of video $l$, and $\textbf{y}_l$ is the corresponding emotion label.
For sample $\textbf{x}_l^{V}$, suppose the feature map of the conv5 in 3D ResNet-101 is $\textbf{F}_l^{V}\in \mathbb{R}^{t\times h\times w\times n}$ (we omit $l$ for simplicity in the following),  where $h$ and $w$ are the spatial size (height and width) of the feature map, $n$ is the number of channels, and $t$ is the number of snippets. We reshape $\textbf{F}^{V}$ as
\begin{equation}
{\textbf{F}^V} = \left[ {\begin{array}{*{20}{c}}
{\textbf{f}_{11}^V}&{\textbf{f}_{12}^V}& \cdots &{\textbf{f}_{1m}^V}\\
{\textbf{f}_{21}^V}&{\textbf{f}_{22}^V}& \cdots &{\textbf{f}_{2m}^V}\\
 \vdots & \vdots & \ddots & \vdots \\
{\textbf{f}_{t1}^V}&{\textbf{f}_{t2}^V}& \cdots &{\textbf{f}_{tm}^V}
\end{array}} \right] \in {\mathbb{R}^{t \times m \times n}},
\end{equation}
by flattening the height and width of the original $\textbf{F}^{V}$, where $\textbf{f}_{ij}^{V}\in\mathbb{R}^{n}$ and $m=h\times w$. Here we can consider $\textbf{f}_{ij}^{V}$ as the visual feature of the $j$-th location in the $i$-th super-frame. In the following, we omit the superscript V for simplicity.

\subsection{Visual Spatial Attention Estimation}
\label{ssec:Spatial}
We employ a spatial attention module to automatically explore the different contributions of the regions in super-frames to predict the emotions. Following~\cite{chen2017sca}, we employ a two-layer neural network, \textit{i.e.} a $1\times 1$ convolutional layer followed by a fully-connected layer with a softmax function to generate the spatial attention distributions over all the super-frame regions. That is, for each ${\textbf{F}_i \in \mathbb{R}^{ m \times n}}(i = 1,2, \cdots ,t)$
\begin{equation}
\begin{aligned}
&{\textbf{H}_i^S} = {{\textbf{W}^{S_1}}({\textbf{W}^{S_2}}{\textbf{F}_i^ \top})^ \top },\\
&\textbf{A}_i^S=\text{Softmax}(\textbf{H}_i^S),\\
\end{aligned}
\label{equ:spatial}
\end{equation}
where $\textbf{W}^{S_1}\in\mathbb{R}^{m\times m}$ and $\textbf{W}^{S_2}\in\mathbb{R}^{1\times n}$ are two learnable parameter matrices, $\top$ is the transpose of a matrix, and $\textbf{A}_i^S \in \mathbb{R}^{ m \times 1}$.

And then we can obtain a weighted feature map based on spatial attention as follows
\begin{equation}
\textbf{F}_i^S=\textbf{A}_{i}^S\otimes\textbf{F}_{i},
\end{equation}
where $\otimes$ is the multiplication of a matrix and a vector, which is performed by multiplying each value in the vector to each column of the matrix.

\subsection{Visual Channel-Wise Attention Estimation}
\label{ssec:ChannelWise}
Assuming that each channel of a feature map in a CNN is a response activation of the corresponding convolutional layer, channel-wise attention can be viewed as a process of selecting semantic attributes~\cite{chen2017sca}. To generate the channel-wise attention, we first transpose $\textbf{F}^{V}$ to $\textbf{G}$

\begin{equation}
\textbf{G} = \left[ {\begin{array}{*{20}{c}}
{{\textbf{g}_{11}}}&{{\textbf{g}_{12}}}& \cdots &{{\textbf{g}_{1n}}}\\
{{\textbf{g}_{21}}}&{{\textbf{g}_{22}}}& \cdots &{{\textbf{g}_{2n}}}\\
 \vdots & \vdots & \ddots & \vdots \\
{{\textbf{g}_{t1}}}&{{\textbf{g}_{t2}}}& \cdots &{{\textbf{g}_{tn}}}
\end{array}} \right] \in {\mathbb{R}^{t \times n \times m}},
\end{equation}
where $\textbf{g}_{ij}\in \mathbb{R}^m$ represents the $j$-th channel in the $i$-th super-frame of the feature map $\textbf{G}$. The channel-wise attention for ${\textbf{G}_i \in \mathbb{R}^{n \times m}}(i = 1,2, \cdots ,t)$ is defined as
\begin{equation}
\begin{aligned}
&{\textbf{H}_i^C} = {\textbf{W}^{C_1}}{({\textbf{W}^{C_2}}{\textbf{G}_i^ \top})^ \top },\\
&\textbf{A}_i^C=\text{Softmax}(\textbf{H}_i^C),\\
\end{aligned}
\end{equation}
where $\textbf{W}^{C_1}\in\mathbb{R}^{n\times n}$ and $\textbf{W}^{C_2}\in\mathbb{R}^{1\times m}$ are two learnable parameter matrices, and $\textbf{A}_i^C \in \mathbb{R}^{ n \times 1}$.

And then a weighted feature map based on channel-wise attention is computed as follows
\begin{equation}
\textbf{G}_i^C=\textbf{A}_{i}^C\otimes\textbf{G}_{i},
\end{equation}
where $\otimes$ is the multiplication of a matrix and a vector.



\subsection{Visual Temporal Attention Estimation}
For a video, the discriminability of each frame to recognize emotions is obviously different. Only some keyframes contain discriminative information, while the others only provide the background and context information~\cite{song2017end}. Based on such observations, we design a temporal attention sub-network to automatically focus on the important segments that contain keyframes. To generate the temporal attention, we first apply spatial average pooling to $\textbf{G}^C$ and reshape it to $\textbf{P}$
\begin{equation}
\textbf{P} = [\begin{array}{*{20}{c}}
{{\textbf{p}_1},} \cdots {,{\textbf{p}_t}}
\end{array}] \in {\mathbb{R}^{t \times n}},
\end{equation}
where $\textbf{p}_j \in \mathbb{R}^{n}(j = 1,2, \cdots ,t)$. Here we can consider $\textbf{p}_j$ as the visual feature of the $j$-th super-frame. The temporal attention is then defined as
\begin{equation}
\begin{aligned}
&{\textbf{H}^T} = {\textbf{W}^{T_1}}{({\textbf{W}^{T_2}}{\textbf{P}^ \top})^ \top },\\
&\textbf{A}^T=\text{ReLU}(\textbf{H}^T),\\
\end{aligned}
\end{equation}
where $\textbf{W}^{T_1}\in\mathbb{R}^{t\times t}$ and $\textbf{W}^{T_2}\in\mathbb{R}^{1\times n}$ are two learnable parameter matrices, and $\textbf{A}^T \in \mathbb{R}^{ t \times 1}$. Following ~\cite{song2017end}, we use ReLU (Rectified Linear Units) as the activation function here for its better convergence performance. The final visual embedding is the weighted sum of all the super-frames
\begin{equation}
{\textbf{E}^V} = \sum\limits_{j = 1}^t {{\textbf{p}_j} \cdot \textbf{A}_j^T}\in \mathbb{R}^{n}.
\end{equation}

\subsection{Audio Representation Extraction}
Audio features are complementary to visual features, because they contain information of another modality. In our problem, we choose to use the most well-known audio representation: the mel-frequency cepstral coefficients (MFCC). Suppose we are given $N$ audio training samples $\{(\textbf{x}_l^{A},\textbf{y}_l)\}_{l=1}^{N}$, where $\textbf{x}_l^{A}$ is a descriptor from the entire soundtrack of the video $V_l$ and $\textbf{y}_l$ is the corresponding emotion label. We center-crop $\textbf{x}_l^{A}$ to a fixed length of $q$ to get ${\textbf{x}_l^{A}}'$, and pad itself when it is necessary. Similar to the method we take in extracting visual representation, we divide each descriptor into $t$ segments and use 2D ResNet-18 ~\cite{he2016deep} as backbone of the audio stream of our model which processes descriptor segments independently. For descriptor $\textbf{x}_l^{A}$, suppose the feature map of the conv5 in 2D ResNet-18 is $\textbf{F}_l^{A}\in \mathbb{R}^{t\times h'\times w' \times n'}$ (we omit $l$ for simplicity in the following), where $h'$ and $w'$ are the height and width of the feature map, $n'$ is the number of channels, and $t$ is the number of segments. We apply spatial average pooling to $\textbf{F}^{A}$ and obtain ${\textbf{F}^{A}}' \in \mathbb{R}^{t \times n'}$.

\subsection{Audio Temporal Attention Estimation}
With similar motivation to integrate temporal attention sub-network into the visual stream, we introduce a temporal attention sub-network to explore the influence of audio information in different segments for recognizing emotions as
\begin{equation}
\begin{aligned}
&{{\textbf{H}^A}} = {\textbf{W}^{A1}}{({{\textbf{W}^{A2}}}{{({\textbf{F}^A}')}^ \top})^ \top },\\
&{\textbf{A}^A}=\text{ReLU}({\textbf{H}^A}),\\
\end{aligned}
\end{equation}
where ${\textbf{W}^{A1}}\in\mathbb{R}^{t\times t}$ and ${\textbf{W}^{A2}}\in\mathbb{R}^{1\times n'}$ are two learnable parameter matrices, and ${\textbf{A}^A} \in \mathbb{R}^{ t \times 1}$. The final audio embedding is the weighted sum of all the segments
\begin{equation}
{\textbf{E}^A} = \sum\limits_{j = 1}^t {{{{\textbf{F}_j^A}}'} \cdot {{\textbf{A}_j^A}}} \in \mathbb{R}^{n'}.
\end{equation}

\subsection{Polarity-Consistent Cross-Entropy Loss}
\label{ssec:PCCE}
We concatenate $\textbf{E}^V$ and $\textbf{E}^A$ to obtain an aggregated semantic vector $\textbf{E} = [\textbf{E}^V, \textbf{E}^A]$ , which can be viewed as the final representation of a video and is fed into a fully connected layer to predict the emotion labels. The traditional cross-entropy loss is defined as
\begin{equation}
{\mathcal{L}_{CE}} =  - \frac{1}{N}\sum\limits_{i = 1}^N {\sum\limits_{c = 1}^C {\mathds{1}_{[c=y_i]}\log p_{i, c}} },
\label{equ:ceLoss}
\end{equation}
where $C$ is the number of emotion classes ($C=8$ for VideoEmotion-8 and $C=6$ for Ekman-6 in this paper), $\mathds{1}_{[c=y_i]}$ is a binary indicator, and $p_{i, c}$ is the predicted probability that video $i$ belongs to class $c$.


Directly optimizing the cross-entropy loss in Eq.~(\ref{equ:ceLoss}) can lead some videos to be incorrectly classified into categories that have opposite polarity.
In this paper, we design a novel polarity-consistent cross-entorpy (PCCE) loss to guide the attention generation. That is, the penalty of the predictions that have opposite polarity to the ground truth is increased. The PCCE loss is defined as
\begin{equation}
{\mathcal{L}_{PCCE}} =  - \frac{1}{N}\sum\limits_{i = 1}^N  (1 + \lambda (g({\hat{y}_i, y_i}))) {\sum\limits_{c = 1}^C {\mathds{1}_{[c=y_i]}\log {p_{i,c}}} },
\label{equ:pcceLoss}
\end{equation}
where $\lambda$ is a penalty coefficient that controls the penalty extent. Similar to the indicator function, g(.,.) represents whether to add the penalty or not and is defined as
\begin{equation}
g(\hat{y},y)=\begin{cases}
1, & \text{if } \text{polarity}(\hat{y})\neq \text{polarity}(y),\\
0, & \text{otherwise},
\end{cases}
\label{equ:indicator}
\end{equation}
where $\text{polarity}(.)$ is a function that maps an emotion category to its polarity (positive or negative). Since the derivatives with respect to all parameters can be computed, we can train the proposed VAANet effectively in an end-to-end manner using off-the-shelf optimizer to minimize the loss function in Eq.~(\ref{equ:pcceLoss}).

\section{Experiments}
\label{sec:Experiments}

In this section, we evaluate the proposed VAANet model on emotion recognition in user-generated videos. We first introduce the employed benchmarks, compared baselines, and implementation details. And then we report and analyze the major results together with some empirical analysis.

\subsection{Experimental Settings}
\label{ssec:Settings}

\subsubsection{Benchmarks}
\label{sssec:Benchmarks}
We evaluate the performances of the proposed method on two publicly available datasets that contain emotion labels in user-generated videos: VideoEmotion-8~\cite{jiang2014predicting} and Ekman-6~\cite{xu2018heterogeneous}.

VideoEmotion-8~\cite{jiang2014predicting} (VE-8) consists of 1,101 videos collected from Youtube and Flickr with average duration 107 seconds. The videos are labeled into one of the Plutchik's eight basic categories~\cite{plutchik1980emotion}: negative \textit{anger}, \textit{disgust}, \textit{fear}, \textit{sadness} and positive \textit{anticipation}, \textit{joy}, \textit{surprise}, \textit{trust}. In each category, there are at least 100 videos. Ekman-6~\cite{xu2018heterogeneous} (E-6) contains 1,637 videos also collected from Youtube and Flickr. The average duration is 112 seconds. The videos are labeled with Ekman's six emotion categories~\cite{ekman1992argument}, \textit{i.e.}  negative \textit{anger}, \textit{disgust}, \textit{fear}, \textit{sadness} and positive \textit{joy}, \textit{surprise}.

\subsubsection{Baselines}
\label{sssec:Baselines}
To compare VAANet with the state-of-the-art approaches for video emotion recognition, we select the following methods as baselines: (1) SentiBank~\cite{borth2013sentibank}, (2) Enhanced Multimodal Deep
Bolzmann Machine (E-MDBM)~\cite{pang2015deep}, (3)  Image Transfer Encoding (ITE)~\cite{xu2018heterogeneous}, (4) Visual+Audio+Attribute (V.+Au.+At.)~\cite{jiang2014predicting}, (5) Context Fusion Net (CFN)~\cite{chen2016emotion}, (6) V.+Au.+At.+E-MDBM~\cite{pang2015deep}, (7) Kernelized and Kernelized+SentiBank~\cite{zhang2018recognition}.

\subsubsection{Implementation Details}
\label{sssec:Details}
Following~\cite{jiang2014predicting,zhang2018recognition}, the experiments on VE-8 are conducted 10 runs. In each run, we randomly select 2/3 of the data from each category for training and the rest for testing. We report the average classification accuracy of the 10 runs. For E-6, we employ the split provided by the dataset, \textit{i.e.} 819 videos for training and 818 for testing. The classification accuracy on the test set is evaluated. Our model is based on two state-of-the-art CNN architectures: 2D ResNet-18 ~\cite{he2016deep} and 3D ResNet-101 ~\cite{hara2018can}, which are initialized with the weights pre-trained on ImageNet~\cite{deng2009imagenet} and Kinetics~\cite{carreira2017quo}, respectively. In addition, for the visual stream, we divide the input video into 10 segments and sample 16 successive frames from each of them. We resize each frame of the visual sample and make the short side length of the sample equal to 112 pixels, and then apply random horizontal flips and crop a random 112 x 112 patch as data augmentation to reduce overfitting. In our training, Adam~\cite{kingma2014adam} is adopted to automatically adjust the learning rate during optimization, with the initial learning rate set to 0.0002 and the model is trained with batch-size 32 for 150 epochs. Our model is implemented using PyTorch.

\begin{table*}[!t]
\centering
\caption{Comparison between the proposed VAANet and several state-of-the-art methods on the VE-8 dataset, where `Visual', `Audio', and `Attribute' indicate whether corresponding features are used, `Auxiliary' means whether no auxiliary data is used except the commonly used ImageNet~\cite{deng2009imagenet} and Kinetics~\cite{kay2017kinetics} for pre-training, and `End-to-end' indicates whether the corresponding algorithm is trained in an end-to-end manner. The best method is emphasized in bold. Our method achieves the best results, outperforming the state-of-the-art approaches.}
\begin{tabular}{ccccccc}
\toprule
Method & Visual & Audio & Attribute & Auxiliary & End-to-end  & Accuracy\\
\hline                                               SentiBank~\cite{borth2013sentibank}              &  \checkmark &       &   \checkmark &   &  & 35.5       \\
E-MDBM~\cite{pang2015deep}                       &   \checkmark     & \checkmark & & \checkmark &  & 40.4        \\
ITE~\cite{xu2018heterogeneous}                   &  \checkmark      & \checkmark & \checkmark  &  & &       44.7        \\
V.+Au.+At.~\cite{jiang2014predicting}            &    \checkmark    &   \checkmark  &  \checkmark &          &  &       46.1        \\
CFN~\cite{chen2016emotion}                       &   \checkmark     &  \checkmark  & \checkmark &  &  &      50.4         \\
V.+Au.+At.+E-MDBM~\cite{pang2015deep}            &    \checkmark    &   \checkmark    &     \checkmark      &       &    &       51.1        \\
Kernelized~\cite{zhang2018recognition}           &    \checkmark    &       &           &     \checkmark    &  &      49.7         \\
Kernelized+SentiBank~\cite{zhang2018recognition} &    \checkmark    &       &  \checkmark         &           &   &    52.5       \\
VAANet~(Ours)                                    &   \checkmark    &\checkmark   &           &    \checkmark       &  \checkmark     &   \textbf{54.5}       \\
\bottomrule
\end{tabular}
\label{tab:SOTA_VideoEmotion}
\end{table*}


\begin{table}[!t]
\centering
\caption{Comparison between the proposed VAANet and several state-of-the-art methods on the E-6 dataset. The best method is emphasized in bold. Our method performs better than the state-of-the-art approaches.}
\begin{tabular}{c c}
\toprule
Method & Accuracy\\
\hline
ITE~\cite{xu2018heterogeneous}                   &     51.2    \\
CFN~\cite{chen2016emotion}                       &    51.8     \\
Kernelized~\cite{zhang2018recognition}           &    54.4     \\
VAANet~(Ours)                                    &     \textbf{55.3}    \\
\bottomrule
\end{tabular}
\label{tab:SOTA_Ekman}
\end{table}


\subsection{Comparison with the State-of-the-art}
\label{ssec:Comparison}

The extracted features, training strategies,  and average performance comparisons between the proposed VAANet and the state-of-the-art approaches are shown in Tables~\ref{tab:SOTA_VideoEmotion} and \ref{tab:SOTA_Ekman} on VE-8 and E-6 datsets, respectively. From the results, we have the following observations:

\textbf{(1)} All these methods consider visual features, which is reasonable since the visual content in videos is the most direct way to evoke emotions. Further, existing methods all employ the traditional shallow learning pipeline, which indicates that the corresponding algorithms are trained step by step instead of end-to-end.

\textbf{(2)} Most previous methods extract attribute features. It is demonstrated that attributes indeed contribute to the emotion recognition task~\cite{chen2016emotion}. However, this requires some auxiliary data to train attribute classifiers. For example, though highly related to emotions, the adjective noun pairs obtained by SentiBank are trained on the VSO dataset~\cite{borth2013sentibank}. Besides high computation cost, the auxiliary data to train such attribute classifiers are often not available in real applications.

\textbf{(3)} Without extracting attribute features or requiring auxiliary data, the proposed VAANet is the only end-to-end model and achieves the best emotion recognition accuracy. Compared with the reported state-of-the-art results, \textit{i.e.}  Kernelized+SentiBank~\cite{zhang2018recognition} on VE-8 and Kernelized~\cite{zhang2018recognition} on E-6, VAANet can respectively obtain 2\% and 0.9\% performance gains. The performance improvements benefit from the advantages of VAANet. First, the various attentions enable the network to focus on discriminative key segments, spatial context, and channel interdependency. Second, the novel PCCE loss considers the polarity-emotion hierarchy constraint, \textit{i.e.} the emotion correlations, which can guide the detailed learning process. Third, the visual features extracted by 3D ResNet-101 can model the temporal correlation of the adjacent frames in a given segment.

\begin{figure*}[!t]
\begin{center}
\centering \includegraphics[width=1.0\linewidth]{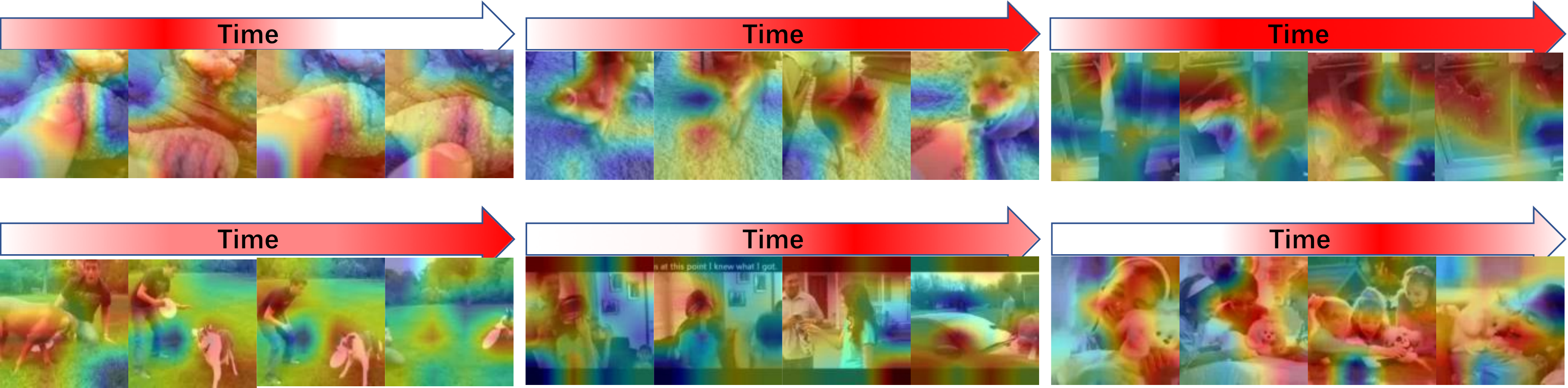}
\caption{Visualization of the learned visual spatial attention and visual temporal attention. In both learned color bar and attention maps, red regions indicate more attention. The proposed VAANet can focus on the salient and discriminative frames and regions for emotion recognition in user-generated videos. Note that all the shown examples are drawn from the test set of VE-8. }
\label{fig:heatmap}
\end{center}
\end{figure*}

\begin{center}
\begin{table*}[!t]
\centering
\caption{Ablation study of different attentions in the proposed VAANet for video emotion recognition on the VE-8 dataset, where `VS', `VCW', `VT', and `AT' are short for visual spatial, visual channel-wise, visual temporal, and audio temporal attentions, respectively. All the attentions contribute to the emotion regression task.}
\begin{tabular}{cccccccccc}
\toprule
Attentions & Anger & Anticipation & Disgust & Fear & Joy & Sadness & Surprise & Trust & \textbf{Average} \\
\hline
AT           &   11.3    &      3.0        &    19.1     &      46.1&  46.1   &    42.4   & 74.7        &10.7       &     \textbf{41.4}    \\
VS           &  45.4   &    27.9          &     44.8    &  59.5    &    49.1 &51.0         &    65.1      &     35.6  &     \textbf{51.7}    \\
VS+VCW       &   46.2    &      25.9        &   42.8      & 63.2     & 50.9    &    40.2     & 67.5         &  45.2     &  \textbf{52.6}       \\
VS+VCW+VT   &   55.6    &      30.8        &   37.5      & 60.4     & 57.7    &50.0         &      65.2    & 34.6      & \textbf{53.6}        \\
VS+VCW+VT+AT &   48.2    &      24.1        &   33.3      &     55.9 & 55.9    &    52.5     &      77.1    &  35.6     &       \textbf{54.5}  \\
\bottomrule
\end{tabular}
\label{tab:Ablation_Attention_VideoEmotion}
\end{table*}
\end{center}

\begin{table*}[!t]
\centering
\caption{Ablation study of different attentions in the proposed VAANet for video emotion recognition on the E-6 dataset.}
\begin{tabular}{cccccccc}
\toprule
Attentions & Anger & Disgust & Fear & Joy & Sadness & Surprise & \textbf{Average} \\
\hline
AT           &   32.4   &      14.1        &   38.5      & 28.1     &  63.9   &    46.5     & \textbf{37.2}  \\
VS           &   59.9    &      44.8        &   49.7      &     46.2 &  35.3   &    62.5     & \textbf{50.8}  \\
VS+VCW  &  58.4     &       52.6       &    49.2     &  46.3    &  44.0   &     65.1    &  \textbf{53.4}  \\
VS+VCW+VT    &   57.1    &      53.1        &   48.0      &     57.0 & 38.7    &    65.0     &  \textbf{54.5}  \\
VS+VCW+VT+AT &   55.1    &      50.6        &   45.7      &     53.7 &  50.3   &    68.9     &  \textbf{55.3} \\
\bottomrule
\end{tabular}
\label{tab:Ablation_Attention_Ekman}
\end{table*}

\subsection{Ablation Study}
\label{ssec:Ablation}
The proposed VAANet model contains two major components: a novel attention mechanism and a novel cross-entropy loss. We conduct ablation study to further verify their effectiveness by changing one component and fixing the other. First, using polarity-consistent cross-entropy loss, we investigate the influence of different attentions, including visual spatial (VS), visual channel-wise (VCW), visual temporal (VT), and audio temporal (AT) ones. The emotion recognition accuracy of each emotion category and the average accuracy on VE-8 and E-6 datasets are shown in Table~\ref{tab:Ablation_Attention_VideoEmotion} and \ref{tab:Ablation_Attention_Ekman}, respectively. From the results, we can observe that: (1) visual attentions even only using spatial attention significantly outperform audio attentions (on average more than 10\% improvement), which is understandable because in many videos the audio does not change much; (2) adding each one of them introduces performance gains, which demonstrates that all these attentions contribute to the video emotion recognition task; (3) though not performing well alone, combining audio features with visual features can boost the performance with about 1\% accuracy gains.

\begin{table}[!t]
\centering
\caption{Performance comparison between traditional cross-entropy loss (CE) and our polarity-consistent cross-entropy loss (PCCE) measured by average accuracy.}
\begin{tabular}{cccc}
\toprule
Attentions   & Loss & VE-8 & E-6\\
\hline
\multirow{2}{*}{VS+VCW+VT}    & CE   &   51.9    &  52.0 \\
             & PCCE &   53.6   &   54.5 \\
\hline
\multirow{2}{*}{VS+VCW+VT+AT} & CE &    53.9   &   54.6\\
             & PCCE &   54.5   &   55.3 \\
\bottomrule
\end{tabular}
\label{tab:Ablation_Loss}
\end{table}

Second, we evaluate the effectiveness of the proposed polarity-consistent cross-entropy loss (PCCE) by comparing with traditional cross-entropy loss (CE). Table~\ref{tab:Ablation_Loss} shows the results when visual attentions (VS+VCW+VT) and visual+audio attentions (VS+VCW+VT+AT) are considered. From the results, it is clear that for both settings, PCCE performs better. The performance improvements of PCCE over CE for visual attentions and visual+audio attentions are 1.5\%, 0.6\% and 2.5\%, 0.7\% on the VE-8 and E-6 datasets, respectively. This demonstrates the effectiveness of emotion hierarchy as prior knowledge. This novel loss can also be easily extended to other machine learning tasks if some prior knowledge is available.

\subsection{Visualization}
\label{ssec:Visualization}
In order to show the interpretability of our model, we use the heat map generated by the Gram-Cam algorithm~\cite{selvaraju2017grad} to visualize the visual spatial attention obtained by the proposed VAANet. The visual temporal attention generated by our model is also illustrated through the color bar. As illustrated in Figure~\ref{fig:heatmap}, the well-trained VAANet can successfully pay more attention not only to the discriminative frames, but also to different salient regions in corresponding frames. For example, in the top left test case, the key object that makes people feel `disgust' is a caterpillar, and a man is touching it with his finger. The model assigns the highest temporal attention when the finger is removed, and the caterpillar is completely exposed to the camera. In the bottom left case, our model can focus on the person and the dog during the whole video. Further, when the dog rushes out from the bottom right corner and makes the audience feel `anticipated', the temporal attention becomes larger. In the middle bottom case, our model pays more attention when the `surprise' comes up.



\section{Conclusion}
\label{sec:Conclusion}
In this paper, we have proposed an effective emotion recognition method in user-generated videos based on visual and audio attentions. The developed novel VAANet model consists of a novel attention mechanism and a novel cross-entropy loss, with less auxiliary data used.
By considering various attentions, VAANet can better focus on the discriminative key segments and their key regions. The polarity-consistent cross-entropy loss can guide the attention generation. The extensive experiments conducted on VideoEmotion-8 and Ekman-6 benchmarks demonstrate that VAANet achieves 2.0\% and 0.9\% performance improvements as compared to the best state-of-the-art video emotion recognition approach. In future studies, we plan to extend the VAANet model to both fine-tuned emotion classification and emotion regression tasks. We also aim to investigate attentions that can better concentrate on the key frames in each video segment.
\section{ Acknowledgments}
This work is supported by Berkeley DeepDrive, the National Natural Science Foundation of China (Nos. 61701273, 61876094, U1933114), the Major Project for New Generation of AI Grant (No. 2018AAA010040003), Natural Science Foundation of Tianjin, China (Nos.18JCYBJC15400, 18ZXZNGX00110), and the Open Project Program of the National Laboratory of Pattern Recognition (NLPR).

\bibliographystyle{aaai}
\bibliography{egbib}

\end{document}